# Closing the empirical loop: autonomous AI agents conduct end-to-end research with human participants

Gabrielle Wehr[1], Reuben Rideaux[1,2,3], Amaya J. Fox[4], David R. Lightfoot[1], Jason Tangen[4], Jason B. Mattingley[3,4,5], Shane E. Ehrhardt[1]

[1] Explore Science, Brisbane, Australia
[2] School of Psychology, The University of Sydney, Sydney, Australia
[3] Queensland Brain Institute, The University of Queensland, St Lucia, Australia
[4] School of Psychology, The University of Queensland, St Lucia, Australia
[5] Canadian Institute of Advanced Research (CIFAR), Toronto, Canada

*Funding*: This work was funded by Explore Science. The workflows and algorithms are proprietary to Explore Science.

*Competing Interests*: Authors affiliated with Explore Science are employees of the company.

*Corresponding author:* shane.ehrhardt@explorescience.ai

**Data Availability Statement**

The data and code that support the findings of this study are openly available on GitHub at https://github.com/Explore-Science/Virtuous-Machines-Towards-Artificial-General-Science.

## Summary

Artificial intelligence systems are transforming scientific discovery by accelerating specific research tasks, from protein structure prediction to materials design, yet remain confined to narrow domains requiring substantial human oversight. Exponential growth of scientific literature and increasing domain specialisation constrain researchers' capacity to synthesise knowledge across disciplines and develop unifying theories, motivating exploration of more general-purpose AI systems for science. Here we show that a domain-agnostic, agentic AI system can independently navigate the scientific workflow – from hypothesis generation through data collection to manuscript preparation. The system autonomously designed and executed three psychological studies on visual working memory, mental rotation, and imagery vividness, executed online data collection with 288 participants, developed analysis pipelines through 8-hour+ continuous coding sessions, and produced completed manuscripts. The results demonstrate the capability of AI scientific discovery pipelines to conduct research with theoretical reasoning and methodological rigour comparable to experienced researchers, though with limitations in conceptual nuance and theoretical interpretation. This is a step toward embodied AI that can test hypotheses through real-world experiments, accelerating discovery by autonomously exploring regions of scientific space that human cognitive and resource constraints might otherwise leave unexplored. It raises important questions about the nature of scientific understanding and attribution of scientific credit.

Scientific discovery constitutes an ongoing development of explanatory theories that advance understanding and prediction of reality[1,2]; contributing to technological and social progress. Throughout history, our capacity to formulate and test theories about the universe has evolved from early philosophical inquiry (Plato[3] and Aristotle[4]) into modern scientific investigation. Fundamentally, conjecture and criticism form the basis for scientific advancement, where theories face rigorous testing against empirical evidence and competing explanations[5,6]. Scientists have achieved remarkable breakthroughs through this approach; however, the scientific endeavour operates within inherent human cognitive constraints. Developing better explanatory theories[1,2] requires researchers to synthesise existing knowledge, formulate testable hypotheses, design rigorous experiments, and interpret results within broader theoretical frameworks. As research output and complexity grow exponentially[7,8] (>2.8 million/yr since 2022; ~5.6% growth[9]), researchers face increasing cognitive and practical limitations in their ability to synthesise existing knowledge and discover novel insights[10]. Individual researchers typically report reading ~200 – 300 articles per year, yet many fields now produce thousands annually, suggesting that complete coverage and awareness of disciplinary literature may no longer be practically achievable[11]. Hypothesis generation increasingly occurs within narrowed conceptual spaces constrained by specialised training, while experimental design can suffer from limited cross-disciplinary methodological exposure. Such limitations may affect scientists' capacity to develop unifying explanatory frameworks that transcend disciplinary boundaries and address fundamental questions – the "better explanations" that can drive scientific progress[1,12].

Computational approaches offer the potential to augment human research capabilities, while alleviating some of these limitations within the traditional scientific process. Early theoretical work in chaos theory[13] and computational principles[14] laid essential foundations for algorithmic processing of complex scientific problems. Today, AI applications are transforming scientific discovery across multiple domains from chemistry[15], synthetic biology[16,17], materials science[18–20], mathematics[21,22], and algorithm development[23]; accelerating research productivity and discovery in these areas. For example, AlphaFold has enabled accurate structure prediction for many proteins within hours[17], substantially accelerating downstream research[24]. However, these specialised AI systems operate within narrow scientific domains, requiring significant human expertise and/or explicit programming to solve predefined problems. Modern transformer architectures[25], combined with large text corpora and scaling of computational infrastructure[26], have enabled the development of large language models (LLMs) which demonstrate broad scope versatility, general reasoning, and fluency in scientific discourse across multiple domains[27–30]. Agentic frameworks leverage these capabilities by integrating LLMs within autonomous architectures capable of goal-directed planning, tool use, and environmental feedback[31]. These systems differ fundamentally from previous AI approaches through their capacity to navigate key phases of scientific inquiry within a unified architecture, including hypothesis generation[32–34], experimental design and execution[35–37], manuscript writing[38], and paper evaluation[39].

The first complete agentic pipeline for *in silico* computational research[38] autonomously generated machine learning research from concept to manuscript, with one such article passing peer review for inclusion at a scientific workshop[40]. Current end-to-end agentic scientific discovery frameworks continue to operate primarily in simulated or abstract digital environments[40–45], typically producing outcomes with predictive power but lacking explanation of underlying causal mechanisms. While these predictions can inform subsequent theories, the absence of direct mechanistic insight constrains hypothesis development and generalisable principles[1,46]. Systems that both generate predictions and transparently articulate causal reasoning would significantly advance autonomous scientific discovery by contributing to explanatory frameworks. Embodied AI offers a pathway to address the limitations of previous (purely computational) approaches, as it integrates perception, action, and reasoning into agents within environments that provide direct feedback[47–49]. As a result, the system can test hypotheses via physical experimentation (utilising the foundation provided by online platforms that interact with humans, through to manipulation of robotic appendages[35]) and collect results[1,46] to support or refute proposed explanatory models. Systems have previously developed such potential in a chemical discovery context by automating certain chemical experimentation tasks (e.g., physically mixing reagents and observing results)[50,51], but these systems lacked the capacity for autonomous hypothesis formulation or refinement of understanding based on experimental results.

Here we explore this potential with the goal of developing an autonomous end-to-end scientific discovery pipeline capable of conducting real-world experiments. To this end, we implemented an agentic system incorporating hypothesis generation through experimental design, physical experimental implementation, data analysis, result interpretation, theory refinement, visualisation, and reporting. Completion of a full scientific study required on average 17 hours' processing time and averaged a total marginal cost of ~$114 USD per research project (not including the human participant payments of ~$4,500 USD for the current experiment). We selected cognitive science as the validation domain based on our expertise and the field's established frameworks for remote experimentation. Given pre-validated cognitive paradigms testing visual working memory (VWM)[52], mental rotation[53], and imagery vividness[54]; we tasked the system with completing three lines of research inquiry. Study 1 conducted a controlled intervention, collecting new data; Studies 2–3 analysed the same 288-participant dataset from Study 1 with distinct hypotheses and analyses. Study 1 examined whether VWM precision and mental rotation performance share representational constraints, finding no correlation between individual performance patterns despite both tasks showing expected difficulty effects; attributed to the established "reliability paradox"[55]. Study 2 explored whether imagery vividness influences how previous stimuli bias current perception and memory, finding that individuals with stronger imagery showed no greater carryover effects between trials, challenging theories that imagery and perception rely on common processing mechanisms. Study 3 investigated whether the precision of VWM predicts broader spatial reasoning abilities, finding negligible relationships and suggesting that apparent connections between visual-spatial tasks reflect general cognitive factors rather than specific shared processes. Together, these results demonstrate the feasibility of AI-driven empirical research, and steps toward an embodied AI framework that navigates all the key components of experimental scientific workflows.

## Methods

The end-to-end pipeline includes: (1) a hypothesis formulation engine that identifies potential research questions and testable predictions by searching and validating novelty, breakthrough potential, and feasibility; (2) an experimental protocol engine that designs methodologies, presented as a pre-registration report following Open Science Framework guidelines[56], and includes preliminary power analyses as required; (3) an implementation engine, currently interfaced with platforms for cognitive science; (4) a data analysis engine that designs and executes a transparent processing pipeline, covering raw data cleaning, outlier analysis, statistical testing, and interpretation of outcomes; (5) scientific decision-making, specifically synthesising and analysing experimental outcomes through inference frameworks to determine follow-up experiments and/or studies; (6) a visualisation engine which designs and constructs a set of figures and tables to illustrate results collated across experiments; (7) drafting of a complete manuscript incorporating visualisations and validated citations; (8) 'peer'-style evaluation; and (9) construction of a final formatted manuscript. It achieves a key goal by bridging discovery from *in silico* computational domains, to the real world, enabling the system to conduct empirical testing of hypotheses with experimental interventions on human participants, and to perform detailed analyses of complex, noisy real-world data. While demonstrated here through cognitive psychology experiments, the architecture employs domain-general principles designed to be applicable across diverse scientific fields and achieves a fundamental goal of the emerging 'self-driving-laboratory' paradigm[57,58].

### *Multi-Agent System Architecture*

The system leverages a hierarchical multi-agent architecture[59] to autonomously produce scientific research through the coordinated and collaborative efforts of a task force of specialised AI agents (**Figure 1**). Unlike deterministic pipelines, each agent functions as an autonomous entity capable of receiving inputs, applying domain-specific reasoning, and producing outputs that advance the investigation. Within the dynamically layered network of agent interactions, consisting of '*orchestrators*' (agents with the ability to coordinate and create further sub-agents for themselves) and '*specialists*' (agents excelling in honed skillsets, such as coding, troubleshooting, or review), a single top-level orchestrator (the master agent) coordinates the entire scientific workflow from beginning to end. This structure results in an emergent cascade of expertise and directed attentional flows, key to successfully navigating the unpredictable course of real-world experimentation[59,60].

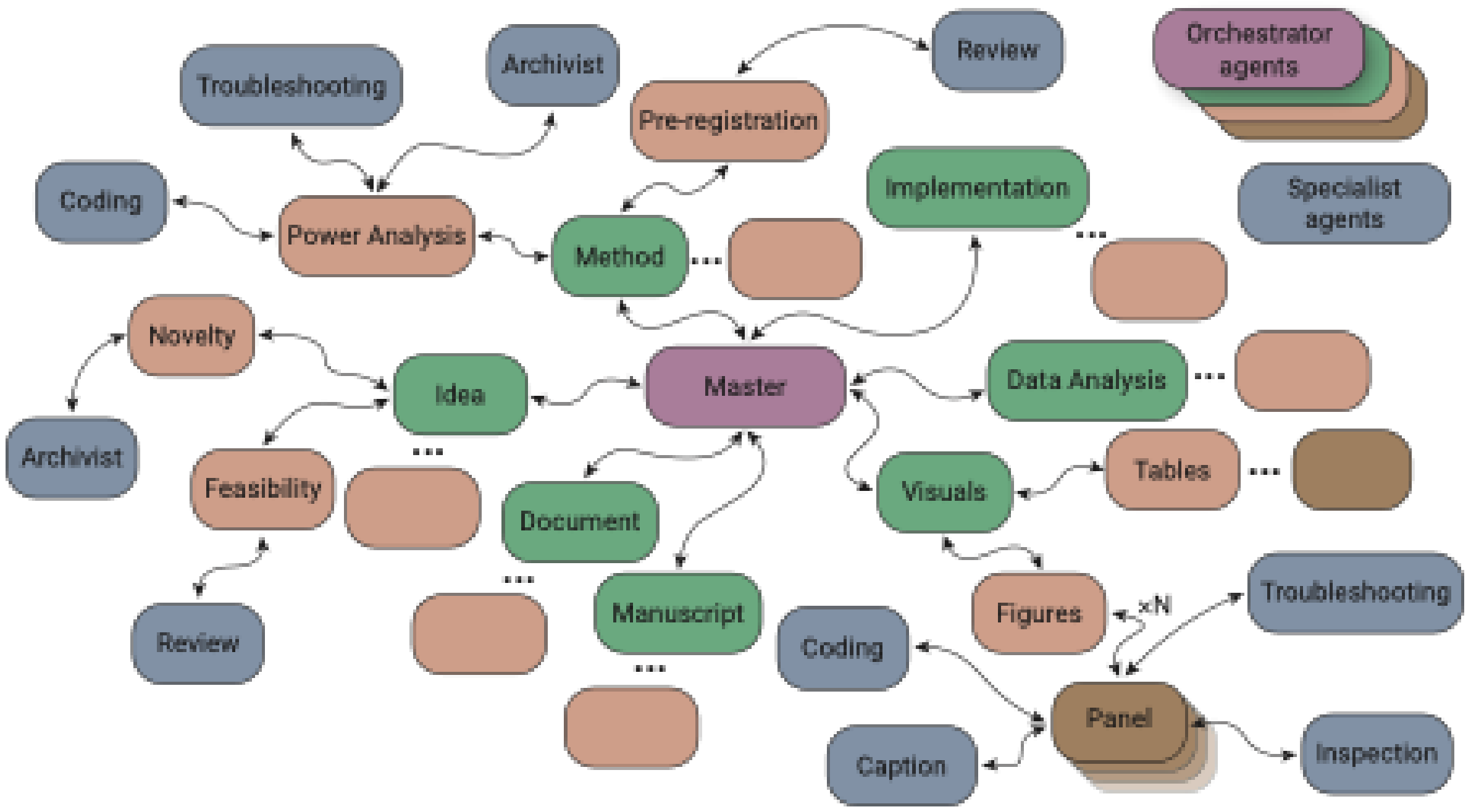

***Figure 1* | Simplified network architecture of the autonomous scientific discovery system.** Directed graph illustrating the information flow and functional relationships between agents. The master agent (purple) coordinates the core scientific workflow agents (green) including method, data analysis, and visuals. Expansion to subagent modules (shades of brown) provides domain-specific capabilities, and interactions with specialist agent pathways (blue) handle specialised tasks including coding, review, troubleshooting, and inspection processes. The stacked panel agent boxes indicate agents completing tasks in parallel, while the dotted connections to blank boxes represent further agent lineages in the system not shown here for clarity. Arrows indicate bidirectional data flow between modules and across hierarchical levels. The distributed architecture enables offloading of complex research tasks while maintaining coherent experimental narratives through the centralised master coordinator.

The system operates as an integrated modular workflow, where the master agent coordinates second-tier orchestrators (e.g., method agent and data analysis agent), each responsible for a specific module of the research project, including idea generation, methodological design, real-world implementation, data analysis, experimental re-evaluation, visuals creation, manuscript preparation, ‘peer’ review, and document construction. Each modular agent functions both autonomously and as a coordinated part of the whole, maintaining independent reasoning threads while remaining responsive to the orchestrated scientific workflow. The orchestrators manage bi-directional information flows with other agents, validating the outputs from each stage to inform subsequent processes with the intention of maintaining methodological consistency, information integrity, and scientific rigour throughout the investigation.

The framework accommodates multiple operation modes along a continuum from autonomous investigation to human-in-the-loop machine learning collaboration models[61]. When running autonomously, the system progresses independently from initial hypothesis to completed manuscript, with each agent making decisions guided by disciplinary standards and outputs from preceding stages. Collaborative modes, on the other hand, allow researchers to provide strategic input at specific intervention points, such as suggesting publications for consideration, bounding decision spaces according to their domain expertise or research priorities, specifying methodological constraints, or outlining visualisation preferences. Researchers can also review and refine all outputs produced by the agents at any stage, providing a means of external quality control throughout the process. The system facilitates these collaborative capabilities without sacrificing its end-to-end functionality, ensuring adaptability across varied research contexts and collaboration models. This flexible architecture enables researchers to deploy the system according to their specific needs – from delegating discrete tasks to commissioning complete investigations with minimal supervision.

### *Human-Inspired Cognitive Operators*

LLMs exhibit broad capabilities[62] yet typically struggle with planning over extended durations and self-verification[63,64]. To address these limitations, we established a foundational cognitive control framework for the system comprising four operators derived from psychological science – abstraction, metacognition, decomposition, and autonomy (**Figure 2**). These operators coordinate planning, tool use, monitoring, evaluation, and refinement across research workflows, serving as computational analogues of the human executive functions that facilitate complex multi-stage inquiries. Each operator draws upon and extends established techniques, combined within the multi-agent system to support empirical investigation with minimal human oversight.

*Abstraction.* The process of focusing on general patterns rather than instance-specific details[65,66] was operationalised as knowledge induction by enabling agents to develop their own heuristics and instructions rather than constraining them with predetermined directives. Concretely, this involved initial elicitation of latent background premises[67], self-driven exploration of problem scope (conceptually related to the 'Self-Ask' method[68]), and automated instruction generation (as validated previously[69]). By beginning with universal principles, the system maintains a broader conceptual search space for potential scientific insights. This implementation mirrors how human scientists maintain conceptual flexibility when developing novel theories[70], allowing exploration across disciplinary boundaries that might otherwise be constrained by specialised training.

*Metacognition.* Awareness and regulation of one's own thinking processes[71,72] was operationalised at two levels, individual and collective, to assess and refine agent reasoning. While frontier LLMs inherently employ forms of internal test-time compute that dynamically scale with task complexity, the system implements explicit self-evaluation protocols that assess evidence quality, logical coherence, and rigour. At the individual level, this was implemented through self-reflective chains of thought[73], enabling each agent to interrogate its underlying assumptions prior to reaching conclusions. At the collective level, agent groups developed awareness of their joint thinking through a reflective process operating on all agents' reasoning traces, similar to 'Tree of Thoughts' inference[74], but across several different agents and utilising an external 'Agent-as-a-Judge'[75] to assess, refine, and arbitrate those traces to align the group on a decided path. These structured self-reflection mechanisms enhance accuracy in complex reasoning tasks[73,76] and facilitate transparent documentation of the evaluation processes.

*Decomposition.* The breaking down of complex problems into more manageable components[77,78] – was operationalised in the framework as explicit structuring of the solution search space. This decomposition enhances the system's capacity to manage the intricacy of multi-stage scientific workflows while maintaining precision at each step. Specifically, parameterisation of logical reasoning steps, conceptually aligned with 'least-to-most' prompting[79], identifies constituent task components. This improves the tractability and transparency of multi-stage scientific workflows and enables verification and refinement of each component to maintain step-level precision. In addition, the system's recursive divide-and-conquer agentic architecture facilitates on-demand subdivision of effort as required, providing flexibility to adapt to challenging tasks[80], thereby improving reliability in the production of the required scientific deliverables.

*Autonomy.* Self-directed goal pursuit[81,82] was implemented in orchestrator agents as local decision-making on tasks, constrained by explicit system objectives. Each orchestrator independently works to complete assigned goals by iterating through a propose-validate-refine process akin to the 'Self-Refine' algorithm[83]. Iteration was governed by three policies: initiation, replanning, and termination. Initiation of sub-agents was invoked upon request of the agent to assist with task completion. Replanning was triggered when validation failed or marginal improvement on acceptance tests fell below a patience threshold. Termination occurred either when validation and quality acceptance checks were all satisfied, or when pre-defined recursion limits were exhausted. Within each iteration, validation of the agents' proposition utilised appropriate external tools where possible to assist with self-correction (shown previously to improve performance[84]). These task feedback signals were then synthesised using a framework analogous to 'Reflexion'[85] and incorporated by the agent to inform subsequent propositions. The iterative self-editing continued until the stopping rules were met, at which point the agent returned its work to its orchestrator.

***Cognitive Offloading and Dynamic Memory***

Humans navigate complex tasks utilising sophisticated memory systems that can: i) hold and manipulate information in working memory[86,87], ii) selectively filter relevant details for the task at hand[88,89], and iii) offload information to external resources when internal capacity is exceeded[90]. For the multi-agent system to maintain coherence over long periods, we emulated these capabilities in the system through the implementation of a dynamic Retrieval-Augmented Generation (d-RAG) system (comparable to ‘DRAGIN’[91]) and construction of retrievable artifacts. Extending existing RAG architectures[92], the d-RAG provides dynamic memory which augments each agent with the cognitive flexibility, prior knowledge, and specific information necessary to carry out its task. Instead of using the same reference material regardless of context, the d-RAG creates and evolves specialised knowledge repositories for each research direction traversed. It functions analogously to how researchers develop domain-specific expertise through targeted literature engagement, and access knowledge during scientific inquiry, combining working memory with cognitive offloading to external resources. The d-RAG forms the core of a multi-tier *'search engine'* tool accessible to the specialist archivist agent, which assists orchestrators requiring real-world information, reducing reliance solely on trained LLM knowledge that may be prone to factual inconsistencies[93]. The search engine facilitates three depths of inquiry: (1) broad academic database searches via the APIs for Semantic Scholar[94], OpenAlex[95], and PubMed, (2) within-text multi-article d-RAG queries (akin to ‘PaperQA-2’[96]), and (3) paper-specific question-answering. The system progressively builds its knowledge base by processing retrieved academic papers in response to agents’ queries, discarding irrelevant retrievals to maintain focused knowledge representations tailored to the specific research question. In addition, by allowing the system to offload complexity to specialised components and file artifacts, a concise and compact representation of the overall research state can be maintained at all points. As evidence of the functional consequences of the d-RAG, ablation of this operation from the system led to production of outputs that neglected recent publications (average most recent = 2019, compared with 2025) while hallucinating others (average hallucinated references = 6, compared with 0). All the outputs produced in the d-RAG ablation test are available on GitHub.

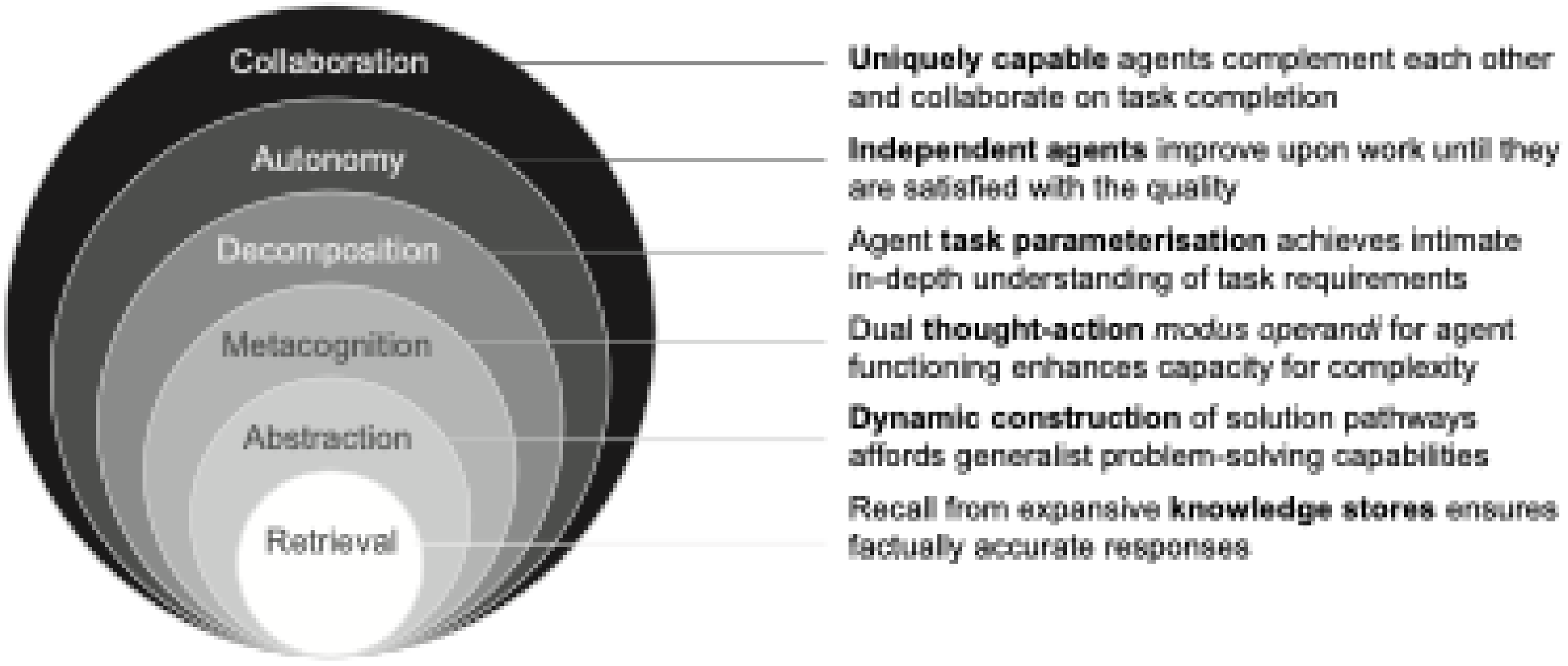

***Figure 2* | Hierarchical framework of cognitive agency levels.** Concentric layers represent ascending levels of agent sophistication, from basic retrieval mechanisms to advanced metacognitive capabilities. Each level encompasses the functionalities of those beneath it while introducing emergent properties. Retrieval forms the foundational layer, providing access to external information and internal

memory stores. Abstraction enables pattern recognition and generalisation beyond specific instances. Metacognition introduces self-monitoring and strategic control of cognitive processes. Decomposition allows complex problems to be systematically partitioned into manageable components. Autonomy confers goal-directed behaviour independent of external guidance. Collaboration represents the highest level, enabling coordinated multi-agent interaction and collective problem-solving.

### *Mixture of Agents*

To increase robustness of the system across the various scientific tasks and unique challenges posed by each, we employed a Mixture of Agents (MoA) approach leveraging complementary strengths of different frontier LLMs[97]. Though MoA may become less necessary as model intelligence advances, current training data choices and reinforcement learning by human feedback processes predispose models to inherent biases that can limit a model's scientific utility. As such, the best performing models for specific aspects of the system were selected to operate conjointly, enabling them to collaboratively accomplish complex tasks that proved highly challenging for any individual model to complete alone. The frontier models utilised include Anthropic's Claude 4 Sonnet, OpenAI's o3-mini & o1, xAI's Grok-3, Mistral's Pixtral Large, and Google's Gemini 2.5 Pro. All models used are standard, publicly available frontier LLMs accessed via their providers' APIs; no privately fine-tuned or otherwise customised model weights are employed anywhere in the system.

### *Core Functional System Components*

#### *Idea Generation*

Frontier LLMs encode high-dimensional representations of knowledge through training on vast text corpora[98], enabling interpolation across distant scientific domains[62] and disparate concepts – such as quantum entanglement and photosynthetic energy transfer – to produce inquiries that may lie in unexplored interstices of existing research. Recent evidence of emergent reasoning in LLMs[99] and their ability to blend conceptual domains[100] supports this notion, suggesting that latent spaces encode abstractions conducive to creative synthesis. The vector-space abstraction enabling transfer in LLMs may be functionally analogous – though not necessarily architecturally identical – to the neural abstraction supporting human task generalisation[101]. Seminal work in psychology suggests transfer between learning contexts requires shared structural elements[102,103] and exemplified the notion of abstraction in cognition[104,105], demonstrating how mental schemas facilitate knowledge transfer across disparate domains. Similarly, LLMs may interpolate between abstracted representations of their training data to produce novel information in new contexts other than the original domains[106,107]. Although LLM-generated ideas are novel in that they are yet to be explored in human documentation, they fundamentally represent an amalgam of current human knowledge and are thus constrained by the models' training data[108]. The system here overcomes this limitation through empirical hypothesis testing, as each completed research idea serves as a basis from which the system can ideate further scientific enquiry (**Figure 3**). By establishing a recursive learning lineage that continually extends its knowledge beyond the original LLM training corpus, it facilitates the potential to yield truly novel hypotheses.

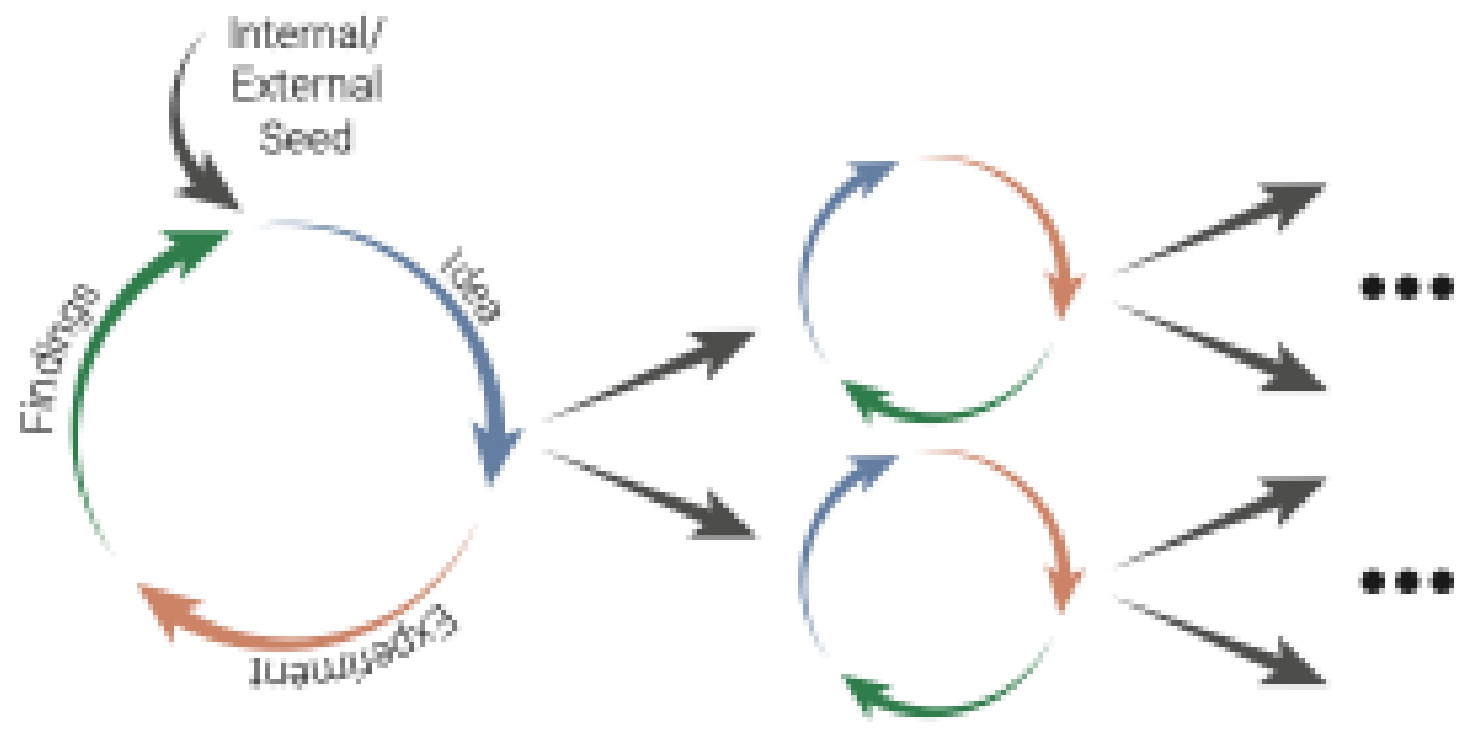

***Figure 3* | Iterative experimentation cycles allow the system to extend beyond trained-on knowledge.** Schematic representation of the self-directed research process executed by the autonomous discovery pipeline. Each circle represents a complete experimental iteration, with transitions between cycles driven by hypothesis refinement and knowledge accumulation. The Internal/External Seed initiates the research trajectory, establishing initial hypotheses or responding to external queries. Progressive cycles demonstrate the capacity for autonomous experimental design, execution, and interpretation, with each iteration informed by previous outcomes. The expanding experimental space explored through successive iterations is characteristic of open-ended scientific discovery.

To perform ideation, the master orchestrator of the system delegates to an idea agent responsible for the formulation of research hypotheses. For the results presented here, we provided seed guidance to the idea agent by specifying the available cognitive tasks and questionnaires, constrained by ethical approval requirements and our domain expertise to ensure validation of each research stage. Such guidance is optional, and if not provided, the agent instead independently targets research fields it considers relevant as the basis for subsequent ideation. The idea agent coordinates several further agents including a review agent which evaluates ideas, a novelty agent which communicates with an archivist agent to perform literature searches via academic databases, and a feasibility agent that assesses methodological constraints and implementation requirements. This multi-agent collaboration aims to ensure that the ideas are scientifically sound, contribute new knowledge, and can be implemented successfully, with ideas passing all checks being constructed into a multifaceted idea framework, which extends that used in previous work[38]. The finalised ideas are ranked through a multi-stage, multi-model tournament process, ultimately identifying a single 'best' idea recommended for further investigation. The full ideation process is summarised in **Figure 4**.

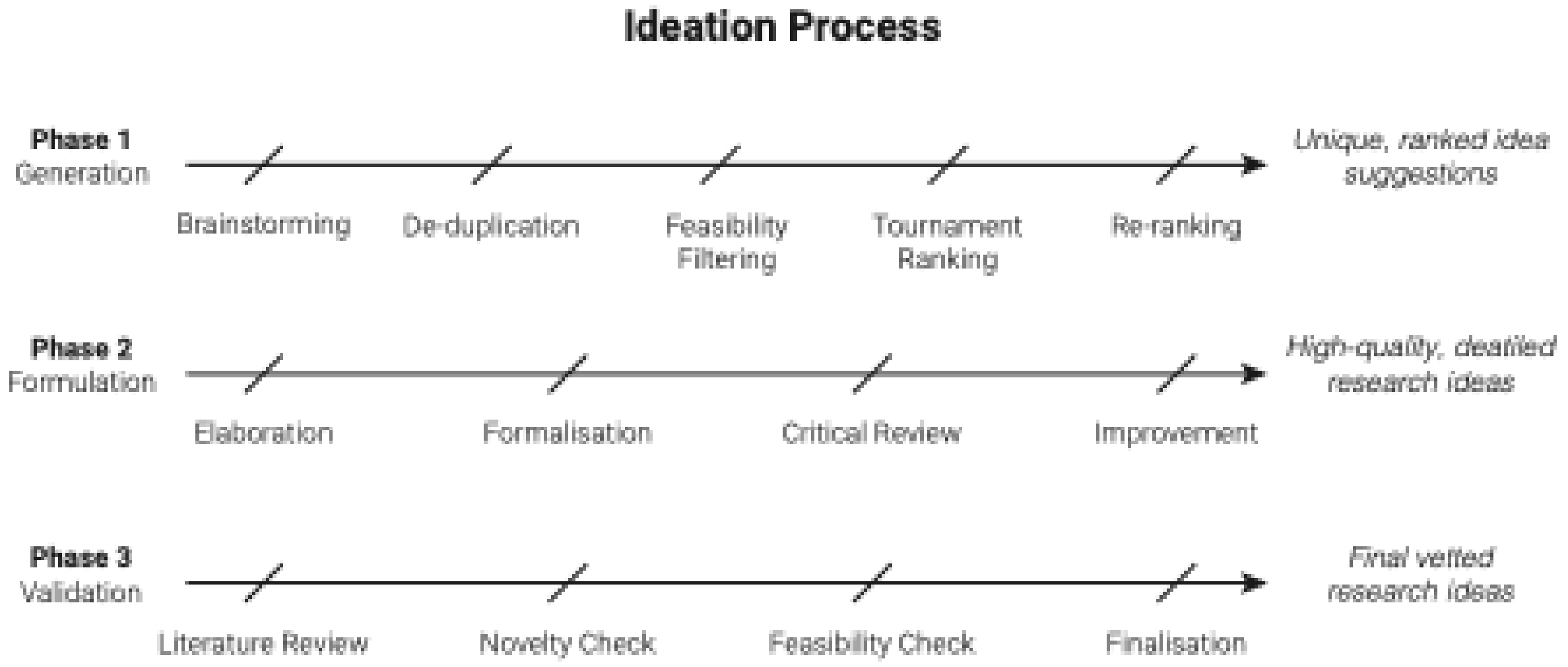

***Figure 4* | Three-phase ideation process for hypothesis generation.** Schematic representation of the iterative scientific ideation workflow implemented to develop novel research questions. Phase 1 (Generation) produces unique, ranked idea suggestions by brainstorming initial concepts, removing redundancies, filtering impractical proposals, ranking by scientific merit, and re-ranking to refine prioritisation. Phase 2 (Formulation) develops the suggestions into detailed research ideas through expansion of conceptual scope, formulation of testable hypotheses, review to interrogate validity, and iterative improvement. Phase 3 (Validation) yields final vetted research ideas via literature review for context, novelty checking against existing work, feasibility assessment for methodological adaptability, and finalisation of a structured research proposal.

### *Methodological Design*

Upon receiving a validated research idea, the master orchestrator delegates it to a method agent responsible for developing an experimental protocol. Through multiple cycles of evaluation with a review agent, a final consolidated plan is optimised for scientific validity, reliability, and robustness. For studies identified as requiring *a priori* statistical sample size determination, the method agent engages a power analysis agent instead of relying on arbitrary conventions. Specifically, the power analysis agent independently conducts the necessary calculations to determine a sample size for the research which satisfies scientific standards while meeting practical constraints. To do so, it coordinates the efforts of several specialists: an archivist agent that queries published literature through the search engine tool to establish the essential parameters for power calculations (including anticipated effect sizes, variance estimates, and appropriate statistical tests); a coding agent that writes and executes power analysis code scripts; and a troubleshooting agent that inspects the validity of the outputs and results to provide feedback. Upon completion, the method agent engages a pre-registration agent, which develops a pre-registration report compliant with Open Science Framework (OSF) standards[56], detailing hypotheses, independent and dependent variables, sampling procedures, exclusion criteria, analytical approaches, and anticipated outcomes. A review agent then evaluates the proposed research plan, corrects any issues found, and revises the power analysis as necessary. This methodological design framework culminates in a finalised experimental protocol that is designed to meet established methodological standards in the field and provide transparency from the earliest stages of the scientific process.

*Real-world Implementation*

For experiments requiring physical-world validation, the master orchestrator engages an implementation agent to manage the execution of the experiment within the confines of the specific tools for which ethics approval has been obtained. For the initial experiment, we focussed on online cognitive psychology paradigms, supporting a previously validated Visual Working Memory Task[52], Mental Rotation Task[53] (MRT), and the Vividness of Visual Imagery Questionnaire-2[54] (VVIQ2). These were delivered to the agent as a single HTML and JavaScript experiment hosted on Pavlovia (GitLab) with integrated participant information, consent, and a SurveyMonkey questionnaire. The system was also interfaced with Prolific – a high-quality, well-established participant recruitment platform for online experiments[109–111] – which is compatible with web-based Pavlovia experiments.

Based on the pre-registration specifications and in alignment with ethical approval requirements, the implementation agent defines and validates the participant recruitment parameters for the Prolific platform – including sample size, age, and vision requirements. In collaboration with a coding agent and troubleshooting agent, code is then written for Prolific's API and executed, which instantiates a complete draft of the study on Prolific (linking to the existing hosted experiment). While the implementation also allows for autonomous live deployment of the draft study via a single call to Prolific's API, for the current work we incorporated a manual verification step and manual publication of the experiment. This additional step, though not technically necessary, enabled careful checking of all parameters before participant recruitment, to ensure human oversight and ethical compliance. Ethics approval for the autonomous studies presented here was obtained from Bellberry Human Research Ethics Committee (HREC; EC00455), after which the draft Prolific experiment for the first study (Study 1) was manually launched for 288 participants (sample size determined by Study 1's power analysis). Two experimenters were present throughout data collection to monitor participant communications. Studies 2 and 3 subsequently developed their own independent hypotheses and analysis plans, calculating optimal sample sizes of 120 and 566 participants respectively. However, all three studies analysed the same 288-participant dataset from Study 1, of which one participant failed to produce any data, and 10 participants failed to complete the study, leaving a final sample of 277 participants with meaningful datasets.

Upon study activation, eligible participants accessed the experimental tasks remotely online through the Pavlovia URL, with participation management data stored on Prolific and experimental response data captured in the GitLab repository. Importantly, all participant data are completely de-identified by the online recruitment platforms to ensure participant privacy; the system never has access to any personal identifying information. Following data collection, the implementation agent inspects the data, analyses its structure, identifies potential quality issues, and prepares documentation of data characteristics for subsequent analysis. This implementation framework provides a generalisable approach to physical-world experimentation that facilitates extension beyond cognitive psychology to other domains requiring human participants or physical interaction.

*Data Analysis*

The master orchestrator passes experimental data onto a dedicated data analysis agent that transforms raw observations into interpretable scientific evidence, prioritising reproducibility, statistical rigour, and explanatory clarity. It manages a multi-stage analytical workflow, collaborating initially with an exclusions agent and archivist agent to establish theoretically informed, literature-backed data cleaning protocols, as well as a validation agent to ensure alignment of the plan with pre-registration commitments. Each analytical stage is subsequently delegated to a coding agent which uses a custom code editing framework based on the file editing functionality of the Aider coding assistant[112], to develop the codebase for the step in a structured environment. Each code execution attempt is evaluated by a team of troubleshooting agents, who offer the coding agent diverse analytical perspectives and feedback on implementation challenges. Progress is documented, providing transparent analytical decision paths. Code refinement continues until the troubleshooting agents and independent verification agents reach consensus that the outputs are satisfactory, or when the predefined iteration limit is reached. The data analysis agent subsequently either proceeds to the next analytical steps until the entire data analysis pipeline has been completed, or requests guidance from the researcher if no further progress is being made. The resulting analytical workflow documents detailed interpretations of the results and the full analytical decision process.

*Experimental Re-evaluation*

Following completion of an experiment, the re-evaluation agent is tasked with determining the next most appropriate step for the research following a structured decision architecture which evaluates all the findings collected up to that point. The decision tree incorporates both Bayesian and frequentist statistical frameworks, where conventional statistical thresholds (Bayes Factor > 10; $p < \alpha$ with sufficient power) function as practical heuristics for the agent to interpret experimental outcomes regardless of the analytical methods employed. Contradictory evidence triggers novel theory generation pathways; theoretically consistent but imprecise results prompt precision enhancement strategies; strong null evidence initiates either theory revision (when conflicting with established frameworks) or alternative hypothesis testing (when consistent with current understanding); and inconclusive evidence prompts study enhancement recommendations. This structured approach formalises scientific judgment, which typically relies on researcher experience and intuition, guiding the progression from initial findings to theoretical understanding, and helping the re-evaluation agent determine whether the study in its current state is complete, or if further experiments are needed for the purposes of theory revision, precision enhancement, or parameter space mapping. The final context-specific recommendation is returned to the master orchestrator to act upon accordingly – either returning process flow back to the method agent if additional follow-on experimentation is required, or noting follow-up study ideas for later use by the idea agent to further traverse the field of research.

*Visual Representation*

Data analysis outcomes for all the completed experiments are handed off by the master orchestrator to the visuals agent, to coordinate the creation of multiple figures and tables essential for clear scientific communication of the findings. The agent formulates a visualisation strategy based on data characteristics and theoretical significance, delegating implementation to two further orchestrators in parallel – a figures agent and a tables agent – each responsible for their respective visual elements. The figures agent instantiates a dedicated panel agent orchestrator for each component of multi-panel figures, establishing a deeply layered reporting structure that maintains cohesion while enabling specialised attention to all individual elements of the visual hierarchy. Each panel agent operates in conjunction with multiple specialists: a coding agent which develops and executes the visualisation code script; a troubleshooting agent that resolves implementation issues; an inspection agent with vision capabilities that evaluates aesthetic clarity; and a caption agent which crafts a detailed figure caption highlighting findings and data representations. The tables agent coordinates an analogous specialist ensemble to transform analytical results into structured tabular formats with accompanying captions, generating tables in both LaTeX and Microsoft Word document formats to accommodate researcher and journal preferences. Through iterative refinement – documented transparently at each stage – these agents collaboratively produce multi-faceted visualisations conforming to disciplinary conventions while attempting to maximise data interpretability and clarity. For methodological visualisations, the visuals agent also modifies provided template scalable vector graphics (SVG) files to represent the specific experimental implementations in the study, employing a caption agent to write figure captions that contextualise the representations within the experimental framework. This approach to visual documentation translates the experimental data and methodology into graphics intended to enhance the accessibility and impact of the research.

*Manuscript Development*

The master orchestrator delegates the process of writing a full research report to a manuscript agent, which leverages the skills of specialists to produce a coherent and high-quality writeup. By engaging an archivist agent, the results are situated within the broader scientific literature, establishing connections with existing theoretical frameworks, and constructing a contextual foundation that enhances the explanatory value of the findings. The manuscript agent also develops a logical scaffold for the scientific narrative, delegating aspects to several specialist writing agents who synthesise all the research components – hypotheses, methodological details, results, visualisations, and theoretical implications – into report sections in collaboration with an archivist agent that provides additional relevant literature references as needed. To address the pervasive tendency of LLMs to generate fictitious citations[113–115], the writing agents utilise a validation tool built on doi.org to verify the authenticity of each reference, enabling correction of fabricated or inaccurate references prior to incorporation in the full manuscript. A review agent provides evaluation of the initial writeup version, guided by domain-specific standards, identifying potential improvements in clarity, logical structure, methodological reporting, and theoretical integration. Based on the feedback, the writing agent incorporates targeted refinements, developing a final manuscript to communicate findings.

*'Peer' Review*

Review of the completed research manuscript is conducted by specialist review agents, following the evaluation protocol used in our domain-agnostic publicly available AI pre-peer review tool Paper Wizard (https://paper-wizard.com). Emulating the process of human peer review, the evaluation examines multiple dimensions of the scientific work, including theoretical foundation, methodological rigour, statistical appropriateness and writing quality. Through detailed assessment, the system identifies both major and minor issues requiring attention in areas where quality could be improved prior to submission for publication. The relevant sections of the manuscript are flagged, and specific, actionable frameworks for addressing each issue are provided, representing a quality assurance mechanism within the system's workflow.

*Document Construction*

The master orchestrator delegates final manuscript assembly to a document agent responsible for incorporating all text, figures, tables, and captions into a single publication-ready file. A multi-format implementation allows researchers to specify a preferred format between LaTeX or Microsoft Word. The LaTeX agent constructs its document in a process that builds upon prior work[38], with modifications to the structure, layout, formatting, and construction of the document, in addition to building out compatibility, user flexibility and robustness. The Microsoft Word agent utilises a similar process to construct its document, with both formats ultimately also being saved to a final PDF format. Several time-consuming technical aspects are handled by the document agents, including placement of figures and tables, section and caption numbering, encoding of mathematical notation and special characters, reference formatting and typographical standardisation, to produce submission-ready manuscripts that adhere to scientific publishing standards.

### *Importance of Feedback*

Central to the system is frequent evaluation, which functions as a key regulatory mechanism integrated throughout the entire research pipeline. The architecture underlying the review process was derived from that of Paper Wizard – the 'peer' review tool – which we tailored to create specialised evaluation protocols for each stage of the scientific process. Feedback allows the system to iteratively refine and adapt its efforts[116,117] until the high standards required for robust scientific inquiry in the field are met. As explicit evaluation and refinement cycles develop higher-quality outputs than single-pass generation[73,83], such intra- and inter-agent feedback mechanisms are fundamental to ensure and maintain methodological rigour and theoretical validity in AI-driven research[118]. However, it is possible that as the intelligence of LLMs continues to advance, the need for extensive review may eventually diminish.

To demonstrate the importance of the review process to the operation of the system, we reran one of the studies presented in the Supplementary Material multiple (3) times after systematically removing the review agents from four stages of the pipeline: methods generation, analysis, visuals generation, and write-up. At a broad level, removing reviewer agents from methods generation, analysis, and write-up caused complete system failure in one of three iterations for each stage, with analysis additionally producing a partial failure (missing sections, tables, or figures) in a second iteration. Outputs generated without visual review fared similarly, with two of three iterations resulting in partial failures. More granular effects emerged across the remaining outputs. Without the methods review agent, fewer validations were performed on model assumptions, such as benchmarking von Mises modelling against alternatives. Without the analysis review agent, sample sizes were substantially reduced, and multiple separate regressions were applied inappropriately across different rotation angles, where a nested approach would have been more suitable. Removing the visuals review agent introduced spatial configuration errors and omitted data from plots. Finally, removing the write-up review agent produced noticeably weaker prose, marked by a less coherent narrative spine, looser sentence construction, and greater internal redundancy, alongside narrower referencing that leaned on a smaller, more repetitive set of sources. All the outputs produced in the reviewer agent removal tests are available on GitHub.

***Safety***

Given the potential for unintended system behaviours when operating with minimal human oversight, the framework incorporates multiple safety measure layers and operational safeguards to ensure system stability, prevent resource overconsumption, and mitigate potential security vulnerabilities. Autonomous code execution is constrained to a timeout dynamically managed by the coding agent, but only up to a ceiling hard limit, preventing excessive runtime while maintaining sufficient flexibility for computationally intensive processes such as statistical modelling. Storage consumption of the code is also continuously monitored and bounded by hard limits to prevent excessive memory use by the system. Package management follows a verification protocol that evaluates each requested library against multiple security criteria: blacklist/whitelist status, typosquatting (slight misspellings of legitimate names) detection, popularity metrics, publisher verification, description analysis, and release history examination. This verification applies recursively to all dependencies, with installation confined to isolated virtual environments to minimise potential global environment contamination. Ubiquitous across all agents, every LLM provider response also undergoes safety evaluation including detection of language pattern shifts, semantic consistency checks, entropy analysis, code syntax validation, and screening for unexpected elements such as arbitrary code blocks, external URLs, or blacklisted keywords unrelated to the scientific task. Additional safeguards include API rate limiting to prevent service overload, logging of all system activities for auditability, and regular checkpointing to enable recovery from potential failures. These nested security measures improve the robustness of the operational framework and enable autonomous scientific discovery while maintaining appropriate boundaries on system capabilities and resource utilisation.

## Results

The aim of this research was to build an end-to-end system capable of producing meaningful scientific knowledge beyond the training data. We deployed a hierarchical framework of AI agents in cognitive science to assess performance on complex tasks and the capacity to coordinate autonomously over extended periods to execute complete scientific workflows. Tasked with independently developing three different studies, the system successfully pursued new lines of scientific inquiry, from conception through analysis and interpretation, leading to empirical findings on human cognition. Beyond manual launch and monitoring of the experiment, downstream stages (e.g., analysis, visualisation, and manuscript drafting) ran autonomously; cosmetic selection of final figure variants for inclusion was performed post hoc. These efforts culminated in complete manuscripts for each study (**Figure 5**; also provided in Supplementary Material 1 – 3), representing the primary result of the work presented here. There are (currently) no objective tests with which to evaluate the quality of scientific manuscripts; thus, here we employed structured expert review (mirroring established peer review processes) to assess the AI-generated manuscripts.

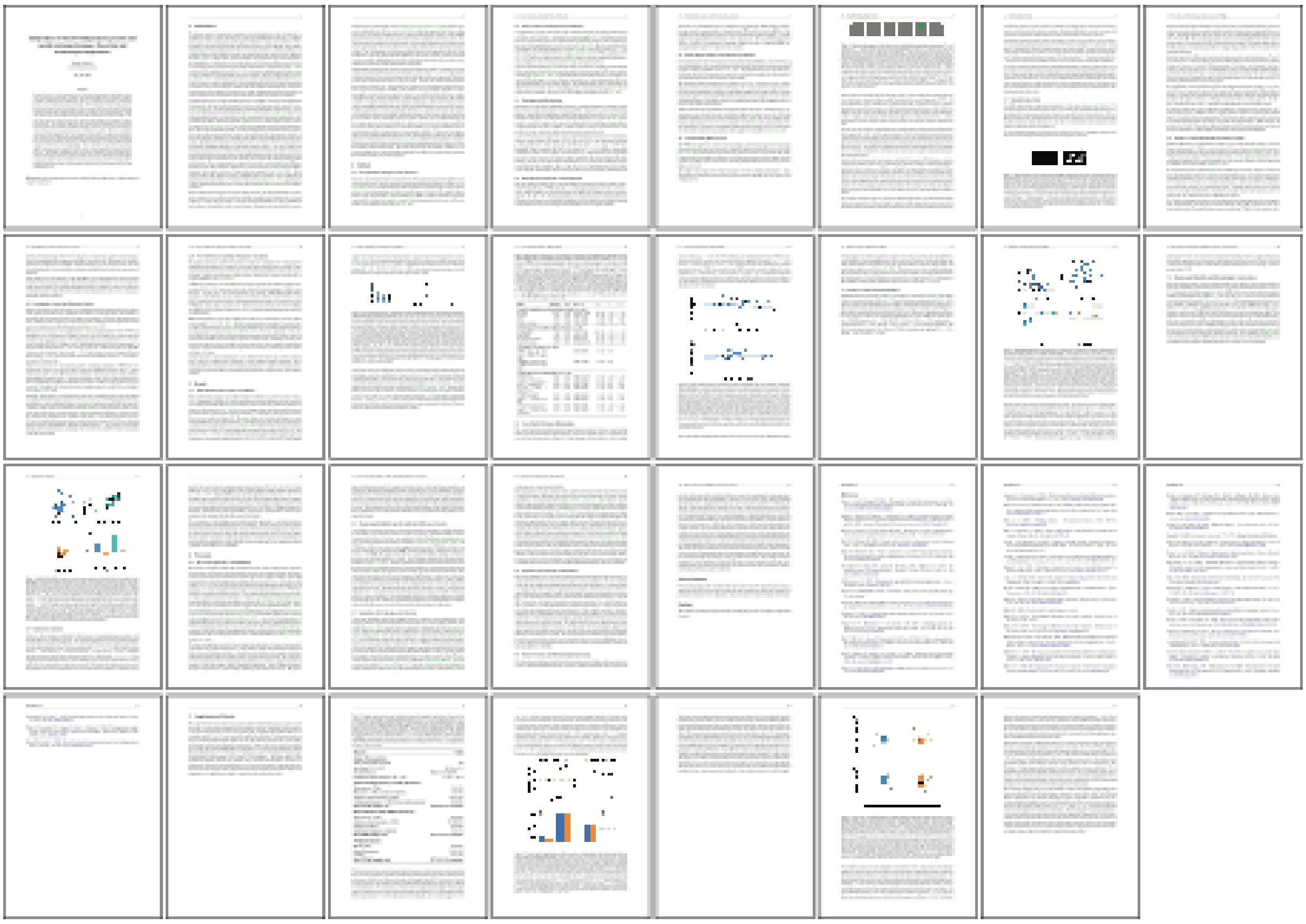

***Figure 5* | Manuscript generated by the pipeline.** The full 31-page manuscript produced in Study 1 spans hypothesis formulation to final formatting and follows a standard scientific manuscript structure with embedded figures (visible in pages 6, 7, 11, 13, 15, 17, 28, 30), tables (pages 12 & 27), statistical outputs, and references. All content was autonomously generated and typeset in LaTeX. This represents one of the three independently generated manuscripts.

### *System Performance*

The system executed full studies (from initial conception to manuscript) in ~17 hours runtime on average per study, excluding data collection. Comparatively, typical human-led workflows can require weeks to months of human expert time depending on experimental complexity. More than 50 agents contributed at various stages, coordinating to execute distinct components of the scientific workflow. Computational usage per study varied with research complexity and difficulty, averaging 32.5 million tokens (distribution detailed in Table 1). In terms of system capabilities, ~1000 – 3000 scientific publications were examined during the literature review process per study. The data analysis agents successfully implemented mixed-effects models and multi-level modelling; handled 279 heterogeneous raw data CSV files; and created 14 – 23 derived CSV files for statistical analysis (all available on GitHub). These agents also showed temporal persistence without human intervention (mean runtime: 8h 32m; SD = 3h 22m), exhibiting goal-directed behaviour to successfully generate functional debugged statistical code. The generated code totalled a mean of 7696 lines (SD = 2426) per study. Each data analysis pipeline involved navigating 72 action-observation cycles on average (SD = 17), including error-recovery and deliberate code-optimisation to meet the specification. Visualisation outputs comprised ~10 – 20 figure panels per manuscript, while each manuscript totalled ~7000 – 8000 words and incorporated ~40 – 50 verified references.

***Table 1* | Token usage and duration for each module of the framework**. Cumulative costs and tokens for all LLMs utilised are shown. Values reported as mean ± SD.

| | Input Token Use (millions) | Output Token Use (millions) | Total Token Use (millions) | Time On Task | Cost (USD) |
|---|---|---|---|---|---|
| Idea | 1.101 ± 0.297 | 0.228 ± 0.066 | 1.330 ± 0.363 | 1h 58m 31s ± 9m 19s | 6.691 ± 1.88 |
| Method | 0.343 ± 0.038 | 0.061 ± 0.004 | 0.404 ± 0.041 | 17m 5s ± 2m 9s | 1.907 ± 0.175 |
| Implementation | 0.021 ± 0.006 | 0.001 | 0.022 ± 0.006 | 30s ± 6s | 0.077 ± 0.017 |
| Analysis | 15.657 ± 7.688 | 0.84 ± 0.34 | 16.497 ± 7.997 | 8h 32m 16s ± 3h 22m 8s | 52.608 ± 29.592 |
| Evaluation | 0.122 ± 0.133 | 0.02 ± 0.025 | 0.142 ± 0.158 | 6m 6s ± 7m 13s | 0.656 ± 0.755 |
| Figures | 7.232 ± 0.813 | 0.871 ± 0.083 | 8.103 ± 0.867 | 3h 45m 52s ± 2h 11m 21s | 29.164 ± 2.604 |
| Writing | 3.605 ± 0.555 | 0.194 ± 0.01 | 3.799 ± 0.564 | 1h 44m 56s ± 53m 39s | 15.539 ± 1.733 |
| Review | 0.586 ± 0.073 | 0.083 ± 0.015 | 0.669 ± 0.087 | 12m 12s ± 1m 39s | 2.183 ± 0.107 |
| Documents | 1.518 ± 0.127 | 0.049 ± 0.005 | 1.567 ± 0.129 | 17m 3s ± 1m 25s | 5.294 ± 0.415 |
| **Total** | **30.184 ± 7.759** | **2.347 ± 0.358** | **32.533 ± 8.067** | **16h 54m 31s ± 4h 7m 15s** | **114.119 ± 29.829** |

### *Human Expert Evaluation*

Expert evaluation by human scientists of the three AI-generated manuscripts identified both strengths and limitations across multiple dimensions of scientific quality. Specifically, while the AI system demonstrated competent use of advanced methods and statistics, as well as comprehensive literature integration; it also exhibited occasional issues with theoretical distinctions, statistical reporting, and interpretation. This mixed competency profile provides insight into the current capabilities and constraints of AI-assisted scientific research, which do not map cleanly onto traditional metrics of research expertise.

#### Positive Aspects

*Clarity and fluency in scientific writing.* All three manuscripts demonstrated clear, professional scientific writing that adhered to disciplinary conventions and maintained coherence across sections. The writing exhibited appropriate use of technical terminology and followed established formatting standards for scientific publication. Study 3's ‘Alternative Mechanisms and Neural Efficiency’ section was particularly well-developed, providing an innovative and theoretically grounded interpretation of the results.

*Creative and theoretically motivated research questions.* The system displayed originality in framing research questions, often exploring relationships not commonly addressed with the given cognitive paradigms. Study 1 explored task order effects and tested multiple model fits to describe the data, demonstrating sophisticated analytical thinking beyond the most obvious research directions. Study 2 investigated the link between mental imagery vividness and serial dependence in VWM and MRT tasks. Study 3 proposed an alternative explanation for null results grounded in individual differences in neural efficiency.

*Rigorous data screening, reliability checks, and control analyses.* Data screening and reliability checks were comprehensive across studies. Study 1 applied split-half reliability analyses, Study 3 checked internal consistency across VVIQ2 subscales, and all studies employed sensible data cleaning protocols with participant- and trial-level screening based on multiple behavioural indicators. These procedures supported data quality and systematically addressed potential confounds.

*Comprehensive and relevant literature engagement.* Literature engagement was thorough and well-integrated into theoretical framing. Study 1 reviewed the historical development of VWM and mental rotation theory, providing detailed context for the research questions. Study 2 noted the "reliability paradox" in individual differences research[55], demonstrating awareness of contemporary methodological discussions. Study 3 effectively contextualised findings in relation to a recent failed replication of foundational results[119].

*Advanced statistical methods and proper error control.* Statistical methods were sophisticated and appropriate across all studies. Mixed-effects models were implemented with systematic comparisons between linear and non-linear fits where relevant. Effect sizes were consistently reported alongside significance tests. Family-wise error correction was properly applied, including Benjamini-Hochberg false discovery rate correction in Study 1.

<u>Negative Aspects</u>

*Theoretical misrepresentations and overstatement.* Theoretical models were occasionally misrepresented, including inappropriate extension of a VWM framework[120] to unrelated visuospatial tasks in Study 3, and Study 2's false claim about working memory resource allocation debates. Study 3 claimed a "novel" imagery-precision link without acknowledging related accuracy research; the statement is misleading given existing literature showing a relationship between imagery strength and VWM accuracy[121]. Studies 2 and 3 also made interpretative claims unsupported by the data.

*Methodological claims and inconsistencies.* Study 1 claimed the VWM cue procedure "isolates memory for the target item" without supporting evidence and described "theoretically unlimited resolution" despite digital constraints. Study 2 extensively discussed measure-reliability but did not explicitly test it (unlike Study 1), leading to unjustifiably confident theoretical interpretations. Study 3 also claimed task engagement metrics validated precision parameter estimates without adequate support for this inference.

*Statistical omissions.* Study 1 did not discuss the incidental significant relationship found between overall VWM error and VVIQ2, evident in Table 1 and Figure 5. Study 2 treated perfectly correlated zero-crossing and width parameters as independent variables, with missing statistics for the width parameter due to this collinearity. Study 3 performed unnecessary multiple correlations across rotation angles, inflating comparisons without justification for expecting effects at specific angles.

*Presentation issues.* Technical problems with visualisations included minor layout issues across Figures, and missing axis labels in Study 3. The figures and tables also often required human selection to identify the most suitable version for inclusion in the manuscript. Other minor problems included inconsistent terminology across Study 1 ("intertrial interval" vs "inter-trial interval"), missing spaces between degree symbols and following words, and awkward table formatting requiring manual adjustments. Study 3 reported "Negligible" instead of actual effect sizes in tables and incorrectly used "sample size" terminology when referring to trial-level data.

*Internal contradictions.* Study 1 contained contradictory statements about methodological advances, advocating for larger samples and more sophisticated statistical approaches while subsequently stating that methodological innovation is needed beyond increased sample size or statistical power. Study 2 made unjustifiably strong claims about multiple comparison corrections addressing problems that "have plagued previous individual differences research" without appropriate qualification. Study 2 also cited different participant numbers in different sections, and Study 3 contained confusing statements about pre-registered thresholds (70%) versus implemented criteria (65%).

*<u>Overall Assessment</u>*

The required use of two cognitive tasks (Visual Working Memory Precision Task and Mental Rotation Task) and a questionnaire on mental imagery (Vividness of Visual Imagery Questionnaire - 2), inherently constrained the range of research questions available to the system. Arguably, the most direct and obvious question – whether individual differences in subjective imagery relate to performance on VWM or MRT – was successfully identified and explored in Study 1. This demonstrates the systems' capacity to recognise theoretically grounded, field-relevant hypotheses. However, it also ventured into less obvious but still interesting territory, such as serial dependence effects, inter-task associations, and exploratory modelling of parameter distributions.

The system particularly excelled in rigour and comprehensiveness. Across the manuscripts, there was a consistent emphasis on transparency, robustness checks, and consideration of alternative explanations. Scientists routinely face interpretative challenges when evaluating statistical outputs, particularly given that sufficiently large sample sizes can yield statistical significance for trivially small effects[122,123] where the meaningful magnitude of difference between data distributions is slight. When faced with statistically significant findings but small effect sizes in Study 3, the system demonstrated notable objectivity and sophistication by prioritising practical significance[124] over statistical significance alone – a commendable departure from the pervasive *p*-value fixation that has long plagued human scientific research[125,126], though it potentially risks undervaluing theoretically meaningful discoveries with modest but reliable effects.

Discussion of prior literature was typically thorough and well-integrated into theoretical framing. However, this strength in systematic reasoning was not always matched by conceptual nuance. The system occasionally struggled to navigate abstract, multi-conditional ideas – particularly those involving subtle theoretical distinctions or long-standing debates within the literature. Interestingly, while the manuscripts often demonstrated cautious and well-qualified interpretations, these were at times juxtaposed with overly broad or confident statements, creating a tension between rigour and over-reach. These weaknesses – including misrepresentation of theoretical frameworks and internal contradictions – are also commonly encountered in human-produced manuscripts, suggesting these limitations may reflect broader challenges in scientific practice rather than being unique to AI-discovery frameworks. In sum, while AI systems can emulate the structure of scientific reasoning, more development is needed for fine-grained judgments that come from deep conceptual familiarity and years of experience navigating complex academic discourse in the field.

*Reproducibility*

The abovementioned assessments address the validity of the research outputs produced by the system; however, they do not assess the reliability of the outputs. In particular, the random seeds used within the system will lead to variations in performance across multiple iterations performed on the same research question and data. To test the reproducibility of the results across multiple iterations, we reran one of the studies (Supplementary Material 3) an additional five times, starting from the stage following research question ideation. While there were considerable differences between the iterations in terms of analytical designs and interpretive focal points, research outputs consistently came to the same broad conclusions across all five runs. All the outputs produced in the reproducibility test are available on the project's GitHub repository.

## Discussion

This study presents an end-to-end AI scientific discovery system that integrates analogues for abstraction, decomposition, metacognition, autonomy, and dynamic memory; and that navigates the complete scientific workflow from hypothesis formulation to manuscript preparation. While domain-specific AI systems have achieved high performance in narrow tasks, such as AlphaFold's protein structure prediction[17], as far as we are aware, this is the first demonstration of an autonomously conducted, end-to-end online experiment with human participants. The results, contained within the three scientific manuscripts that accompany this article, show that AI systems can conduct scientific inquiries with minimal human intervention. While prior systems have primarily operated within computational domains (simulations and modelling), the system's ability to collect and interpret real-world experimental data from human participants marks a key step toward embodied scientific AI that is capable of testing hypotheses in real-world settings. These findings bridge the gap between *in silico* AI capabilities and practical scientific application, with important implications for both the future of AI-augmented scientific discovery and fundamental questions about the nature of knowledge creation.

The contribution of this work is architectural rather than methodological at the model level: we use only standard, publicly available frontier LLMs, and demonstrate that when combined across providers, embedded in a hierarchical multi-agent architecture, and augmented by the cognitive operators and dynamic memory systems described above - these techniques are jointly sufficient to conduct empirical science autonomously from hypothesis generation through to manuscript preparation. To our knowledge, no individual frontier model achieves this end-to-end capability on its own, and no prior agentic framework has demonstrated it with real human-participant data collection.

### *System Performance and Capabilities*

*Efficiency.* The system demonstrated efficiency gains compared with traditional research timelines, executing complete research projects in hours rather than the weeks or months typically required by research teams. At ~$114 USD average marginal cost per complete study (not including human recruitment costs), the system also offers cost reduction compared to even modest empirical studies, which often require substantial investments in researcher salaries, infrastructure, and overhead. The resulting productivity improvements suggest potential for transformative changes in how scientific research is conducted, and the pace at which scientific progress can be made. This cost efficiency could particularly benefit resource-constrained institutions and developing nations, potentially redistributing global scientific capacity.

*Rigour*. Across the three studies, the system tended toward conservative methodological choices, appropriate statistical power, and transparent reporting of limitations – indicating impartiality and cognitive flexibility which can at times challenge human researchers. Importantly, the system's documentation of all analytical decisions and availability of raw data enables reproducibility, which goes some way toward addressing reproducibility concerns in scientific literature where 70% of researchers have failed to reproduce another scientist's experiments[127–129]. The ability to replicate experimental conditions and analyses at negligible marginal cost could alter how we validate scientific findings, and aligns with the growing movement toward open science practices[130].

*Sophistication*. The system demonstrated the capacity to autonomously construct and execute multi-step data analysis pipelines, employing valid statistical techniques, evaluation of underlying assumptions, and measured interpretation of results. The system here sustained coherent coding and statistical reasoning for longer than eight hours – for context, a ~50 – 55-minute 50% task-completion time horizon has been reported for Claude 3.7 Sonnet on research-engineering tasks[131].

*Adaptability*. When encountering unexpected results, the system's problem-solving and decision-making capabilities showed real-time adaptation to outcomes and implementation issues beyond fixed plans. It dynamically modified its approach when confronted with unanticipated outcomes or implementation challenges (e.g., finding appropriate solutions to statistical models not converging), without fixating on the original approach. Coupled with continuous documentation of all decision processes, the resulting audit trail was both comprehensive and transparent.

*Grounding*. By integrating LLMs with the d-RAG to contextualise findings, the system also exhibited quality scientific writing and contextual framing. It often situated findings within broader theoretical contexts, identified appropriate connections with relevant literature, and articulated limitations with clarity. The ability to engage with conceptual and communicative dimensions of scientific discourse suggests effective leveraging of its broad knowledge of relevant published literature.

### *Limitations and Challenges*

Despite its achievements, the system exhibits several limitations that highlight areas for future development. Experimental implementation poses a fundamental bottleneck in the verification of ideas of autonomous science systems[132]. While the framework is general by design, currently its physical capabilities are domain-constrained to online experiments for which the necessary tools and interfaces are available. Representing an engineering challenge rather than a fundamental limitation, the extension of the system to other scientific fields requires the development and incorporation of new and existing domain-specific toolsets, which are readily accommodated by the modular and generalist core system architecture.

Furthermore, while the system was capable of end-to-end research, it was not infallible and can thus benefit from human refinement and oversight. In particular, data visualisations occasionally include graphical imperfections which persist despite inspection and verification mechanisms – such as overlapping axis labels, misaligned label positioning, or suboptimal axis bounds. Though predominantly aesthetic rather than substantively inaccurate, these minor discrepancies highlight the intrinsic challenges in automated computer vision tasks that human visual perception resolves effortlessly[133,134]. Such artifacts remain difficult for the system to detect autonomously, yet can be quickly and easily rectified with minimal human intervention.

In addition, rigorous scientific inquiry involves lengthy, complex, multi-stage processes, requiring the system to handle extremely long chains of thought – thousands of reasoning steps and conversational exchanges extending over 12 hours. The dynamic memory system and cognitive offloading mechanisms were central to maintaining conceptual continuity across research stages, effectively combining working memory with strategic external resource utilisation. The system's ability to selectively filter relevant information while preserving focused knowledge representations of the necessary context enabled it to navigate the entire scientific workflow without conceptual drift. Interestingly, many reasoning models generally degrade in their performance[135,136] over extremely long chains, losing focus and coherence across iterations.

Sensitivity to early-stage accuracy also emerged as a challenge. Poor question formulation or any conceptual errors introduced during hypothesis generation and methodological design propagate downstream – persisting through multiple verification cycles – at a detriment to research outcomes. This potential single-point failure mode likely reflects the “anchoring” bias characterised in LLMs[137], whereby the first logical claims encountered by the model are weighted as ground truth and only weakly revised later, disproportionately shaping subsequent judgements. Consequently, once a misconception enters the chain of thought, later processes tend to build upon it rather than correcting it[138]. Explicit error-checking steps only seldom succeed in reversing the trajectory once a false premise has been internalised. Interestingly, humans also exhibit this cognitive bias[139–141], where initial information acts as a reference point that substantially affects how subsequent information is assessed, particularly under uncertainty. This phenomenon emphasises the importance of robust verification protocols during the earliest conceptual stages of automated scientific investigation and highlights a crucial advantage of human-AI collaborative scientific workflows where human expertise can intervene most effectively at points of highest conceptual leverage.

A related concern is the potential for bias arising from the human-generated corpora on which the underlying large language models were trained. LLMs are known to occasionally produce outputs that align with user or human expectations even when these conflict with reality (for instance, by hallucinating references to non-existent publications), and analogous tendencies in autonomously generated research therefore warrant vigilance. The system mitigates the most overt manifestations through built-in veracity-checking mechanisms (e.g., validation of cited references against external sources), but the broader question of whether the system preferentially generates results that conform to its own prior expectations is less straightforward to address. Two lines of evidence indicate that this is not a concern in the present work. First, the system produces a complete paper-trail of analytic decisions and code, from raw data through to reported results. This pipeline was inspected and re-executed locally to confirm that the analyses were valid and reproduced the values presented in the manuscripts, ruling out the more egregious forms of bias in which invalid or fabricated findings would be generated.

Second, prior to receiving access to the data, the system was required to generate explicit hypotheses, which were recorded in pre-registration documents (available on the project's GitHub repository). If these hypotheses reflect the system's expected outcomes (a reasonable assumption), and if the system were biased toward producing results that conform to those expectations, the pre-registered predictions would be expected to be supported by the subsequent analyses. In practice, the system's hypotheses were frequently violated; for example, in the study titled "Precision Response Patterns in Visual Working Memory: Shared Constraints with Mental

Rotation and the Role of Imagery Vividness", only two of five pre-registered hypotheses (H1a and H1b) received any support, and even these were only partially supported by the data, with similar contradictions between predictions and findings emerging in the other two studies. The force of this pre-registration check derives in part from the fact that the hypotheses were evaluated against experimental data collected specifically for the present studies. Whatever influence the training corpora exerted on the system's priors, those priors could not determine the specific values that would be observed; support for any pre-registered hypothesis therefore had to be derived from measurements the system had not previously encountered. The frequent violation of these hypotheses thus constitutes a substantive empirical test of the system's outputs, rather than a mere procedural formality.

More broadly, the training corpora on which the underlying LLMs were optimised will inevitably shape the system's analytic and interpretive choices, just as a human researcher's prior training shape theirs. Such influences are not inherently problematic, if those influences lead the AI Scientist to produce valid and reliable outputs. Further, and perhaps most importantly, this same issue is present (perhaps even more so) in research conducted by human scientists, where not only the training data (previous experience, papers read, etc.), but also environmental pressures (e.g., job security, academic success, etc.) can influence the types of questions asked, the approach followed in testing them, analytic strategies and theoretical interpretations. In contrast to the AI Scientist, which produces a complete paper-trail of research decisions and actions that can be interrogated, the choices made by human scientists are typically opaque (or at least implicit) and are therefore considerably more difficult to assess and evaluate. Finally, the architecture presented here was sufficient to produce a complete, end-to-end empirical research workflow, but it is not claimed to represent an optimal configuration of agents, tools, or coordination strategies. A growing number of multi-agentic systems and general-purpose agentic frameworks have recently been proposed (e.g., BiOMNI[142]), and it is plausible that some of these could in principle be marshalled toward achieving similar scientific tasks. However, to the best of our knowledge, no existing system has previously demonstrated and reported in the academic literature a fully autonomous end-to-end empirical research workflow that includes data collection from human participants. The novelty of the present contribution therefore lies in the demonstration of this feat, rather than in any claim of architectural exclusivity or superiority over alternative frameworks. The present work is intended as a proof-of-concept, and the underlying components are expected to continue to be refined as the field matures. Principled comparison and optimisation of such systems is currently constrained by the absence of validated instruments for benchmarking the quality of autonomously generated research. Existing evaluation frameworks for AI in science predominantly target narrow, well-defined tasks, such as predictive accuracy on curated datasets, and do not capture the integrated competencies required across hypothesis generation, experimental design, data collection, analysis, interpretation, and reporting. In the absence of such benchmarks, the present study relied on expert human judgement and on releasing the system's outputs in full so that readers may independently appraise their quality. The development of validated, multidimensional benchmarking tools (capable of assessing novelty, methodological rigour, statistical validity, theoretical coherence, and reporting quality) represents an important next step for the field, providing the objective signal required to systematically compare architectures and to optimise the performance and reliability of individual agents and the broader system.

***Future Directions***

Our work points toward several high-impact future developments for autonomous scientific systems. The modular, domain-agnostic architecture we implemented here provides a foundation for general-purpose scientific AI that can operate across research domains and accommodate various methodological techniques and constraints. Expanding beyond cognitive psychology represents an immediate opportunity. Applying the system to fields ranging from medicine to environmental science would require only domain-specific implementation interfaces and minimal changes to the fundamental scientific workflow engine. A promising direction involves integration with physical laboratory automation and robotics, enabling direct manipulation of physical systems across chemistry, biology, and materials science. Enhancing the system's capacity for autonomous theory refinement also represents a key direction for development. Incorporating intrinsic mechanisms for theoretical updating based on empirical results could potentially enable the system to produce more creative scientific contributions and breakthroughs of greater novelty. In addition, continued work on improving cognitive reasoning frameworks is important for strengthening the robustness and quality of research produced – advancing toward a system capable of generating truly novel experimental methodologies and theoretical insights.

Models with real-world experimental capabilities represent a potential pathway toward more advanced AI. The framework we have developed embodies the fundamental cycle of knowledge creation: executing movements to explore the world, reflection on outcomes, understanding relationships with prior knowledge, generating insights from these interactions, and developing foresight for future predictions[143]. This process mirrors cognitive development in children, who learn primarily through physical manipulation of objects and progressive refinement of their understanding[105,144]. While current LLMs excel at pattern recognition within training data, they remain limited by their inability to autonomously expand beyond those boundaries. Here the system coupled internal representations – the exploration of latent connections between concepts – with external measurement through experimentation. This creates a virtuous cycle where hypotheses conceived in the model's rich latent space can be validated against reality, with results feeding back to refine its conceptual framework. This suggests a fundamentally different approach to advancing AI capabilities – one rooted in the power of scientific investigation to build increasingly accurate models of the world. Toward this end, we use the intermediary milestone of ‘Artificial General Science’ (AGS) to denote autonomous systems capable of independently driving scientific inquiry across all domains – generating hypotheses, orchestrating experiments, and iteratively refining knowledge through empirical evidence. In a recursive loop of discovery and learning, the scope of understanding and knowledge of the system could be continually expanded beyond trained-on human knowledge.

***Publishing Autonomously Generated Research***

Evaluating the quality of autonomously generated research through the established channels of academic publishing represents a key open question for the field. Submission to mainstream peer-reviewed venues, however, is presently precluded by the generative-AI policies of all major scientific publishers, which require that submissions either be authored by, or be subject to substantive intellectual contribution from, human researchers. Consequently, fully autonomous research outputs of the kind presented here are currently not eligible to be assessed through standard peer review.

The question of whether AI-generated research should be treated equivalently to human-generated research within the publishing system is non-trivial and warrants ongoing dialogue. One justification for differential treatment is the presumption that human-authored work is subject to greater oversight than autonomously generated work – most directly through the iterative cycles of internal review, co-author feedback, and revision that typically precede submission. This presumption is not always well-founded, however: the quality of human oversight varies considerably across research groups and career stages, and in some circumstances may be poorer than the structured verification mechanisms embedded within autonomous pipelines. Nevertheless, within the current publishing infrastructure, accountability for the content of any submission must ultimately rest with one or more identifiable human authors, who can be held responsible for the integrity of the work and for any subsequent corrections. As autonomous research systems mature, publishers, funders, and learned societies will increasingly need to develop policies that distinguish meaningful human accountability from superficial human attestation, and that articulate the conditions, if any, under which fully autonomous outputs may legitimately enter the scientific record. Dedicated venues such as Agents4Science (https://agents4science.stanford.edu) are likely to provide important early signals on what such evaluation might look like in practice, while broader publishing norms are renegotiated.

***Societal and Ethical Implications***

Autonomous scientific systems capable of designing and executing rigorous methodological plans will likely reshape the role of human scientists[145]. While capable of independent operation with minimal human intervention, the balance between autonomy and human collaboration offers distinct advantages depending on the research goal. Currently, humans provide most value to these systems in creative problem formulation, conceptual innovation, and ethical oversight – though the focal points for human contribution may shift as the technology advances. These systems can also address limitations of contemporary research, particularly the increasingly narrow specialisation that often constrains scientists to their areas of expertise[146], impeding progress on complex interdisciplinary problems. By bridging disciplinary boundaries, domain-agnostic frameworks facilitate integrative research, empowering scientists to explore theoretically adjacent areas where they may lack training but possess valuable conceptual insights. Researchers can delegate any aspect(s) of the scientific workflow to the system as needed, enhancing both efficiency and innovation through collaboration.

The accelerated pace of research enabled by autonomous scientific systems presents both opportunities and challenges. While rapid knowledge generation could expedite solutions to pressing global issues from climate change to disease, inadequate consideration of downstream effects risks unintended consequences. Of particular concern is the potential for producing research outputs at a volume that overwhelms human researchers, necessitating innovative approaches to distil and communicate findings effectively within human cognitive capacity constraints. Notably, autonomous scientific systems are likely to generate a high proportion of null findings, which, while typically remaining unpublished in traditional research[147], offer under-utilised value. By documenting non-significant outcomes, these results may mitigate publication bias[147], as well as characterise the "null space" of research fields – highlighting where relationships are absent and conversely where they may exist. This dual benefit reduces wasted resources on redundant experiments and can guide researchers toward better-informed hypotheses for promising investigations. The environmental impact of running multiple LLMs over sustained periods also warrants consideration, as the material energy and carbon footprint of such systems is non-trivial, and yet to be quantified relative to the net sustainability of human labour. Balancing the dynamics of speed, volume, environmental impact, and meaningful human oversight will be important as autonomous discovery systems become more prevalent in scientific workflows.

The ethical dimensions of these systems are multifaceted, with their potential to democratise high-quality research capabilities representing a promising long-term benefit. By reducing resource barriers, these systems can broaden participation in the scientific enterprise, enabling individuals, institutions and regions with limited infrastructure to contribute to progress on major challenges. However, as their research capabilities expand, safeguarding these systems becomes critical – vulnerabilities to hacking or LLM prompt injection attacks[148] could lead to misuse with potentially severe consequences. Adequate safety mechanisms and optional human verification at each stage of pipelines may help mitigate these risks, yet scalable, comprehensive frameworks will be integral to balancing autonomy with responsible oversight. Beyond safety, ethical considerations include attribution of scientific credit, responsibility for research outcomes, and governance of AI-driven scientific systems. As these systems increasingly contribute to knowledge production, accountability frameworks must adapt to ensure transparency, trust, and fair credit allocation. This is of particular importance in scientific contexts where knowledge claims carry significant societal implications[149,150], whether beneficial or harmful. Delineating responsibility between human researchers and autonomous systems is complex, requiring clear ethical guidelines and governance structures to maintain public trust. The capability to generate thousands of studies rapidly also raises concerns about potential misuse, including automated *p*-hacking, deliberate generation of misleading findings, and high-volume/low-quality outputs that could strain peer review systems. Establishing quality control mechanisms, transparency standards, and detection systems for AI-generated research will be essential as these technologies proliferate. Such measures will enable human scientists to replicate autonomous studies, particularly those with highly impactful findings, to independently validate results, ensure ethical and safety compliance, and facilitate integration into existing societal infrastructure (e.g. patent filing). These ethical implications highlight the need for ongoing dialogue to align AI-driven scientific discovery with principles of safety, equity, and responsibility.

On a philosophical level, the empirical results of this study raises intriguing questions about the nature of knowledge, particularly the role of understanding in knowledge generation. While traditional epistemological frameworks posit human comprehension as an intrinsic component of knowledge creation[151], the system shown here demonstrates that structured scientific inquiry can produce valid empirical knowledge without requiring human-like understanding. Consequently, knowledge may be derived from mechanistic processes without necessitating conscious insight[152,153] – a distinction that invites consideration of how we conceptualise scientific knowledge and the processes through which it emerges.

In sum, here we have presented an AI scientific discovery system which demonstrates that frontier LLMs augmented with human-inspired cognitive operators and physical interaction capabilities can independently execute end-to-end research. In ~17 hours of system runtime (excluding data collection), with minimal human oversight, the system conceived, ran, analysed, and produced complete manuscripts for an online psychology experiment with 288 human participants. We replicated this capability across three distinct studies. As far as we are aware, this is the first demonstration of autonomous, end-to-end experimental research with human participants. Working in collaboration with human researchers, we foresee systems that can accelerate and elevate the rigour of all components of individuals' scientific workflow in the pursuit of high-quality science. Although work remains in enhancing creative theory development and expanding experimental scope, the performance of this system marks a step forward in AI actively participating in empirical investigations of the natural world, through structured scientific inquiry. The epistemological implications of knowledge generation by artificial systems invite reconsideration of traditional frameworks for scientific understanding, and as their capabilities develop further, thoughtful consideration of scientific integrity, safety, and inclusion.